\documentclass[runningheads]{llncs}

\usepackage{cite}
\usepackage{graphicx}
\usepackage{comment}
\usepackage{amsmath,amssymb}
\usepackage{color}
\usepackage{url}
\usepackage{hyperref}

\usepackage{times}
\usepackage{epsfig}
\usepackage{graphicx}
\usepackage{amsmath}
\usepackage{amssymb}
\usepackage{subfig}
\usepackage{tabularx}
\usepackage{multirow}
\usepackage{math}

\usepackage[font=small]{caption}
\usepackage{wrapfig}
\usepackage{array}
\newcolumntype{H}{>{\setbox0=\hbox\bgroup}c<{\egroup}@{}}
\input{author-citation}

\graphicspath{{images/}}

\usepackage[acronym]{glossaries}
\newacronym{har}{HAR}{Human Action Recognition}
\newacronym{gnn}{GNN}{Graph Neural Network}
\newacronym{gcn}{GCN}{Graph Convolutional Network}
\newacronym{cnn}{CNN}{Convolutional Neural Network}
\newacronym{rnn}{RNN}{Recurrent Neural Network}
\newacronym{gft}{GFT}{Graph Fourier Transform}
\newacronym{imu}{IMU}{Inertial Measurement Unit}
\newacronym{mlp}{MLP}{Multilayer Perceptron}
\newacronym{lstm}{LSTM}{Long Short-Term Memory}
\newacronym{gru}{GRU}{Gated Recurrent Unit}
\newacronym{sgd}{SGD}{Stochastic Gradient Descent}
\newacronym{adam}{ADAM}{Adaptive Moment Estimation}
\newacronym{ca}{CA}{Cosine Annealing}
\newacronym{cawr}{CAWR}{Cosine Annealing with Warm Restarts}
\newacronym{nas}{NAS}{Neural Architecture Search}

\newcommand{\andothers}{et al.\,}
\newcommand{\figname}{Figure\ }
\newcommand{\tabname}{Table\ }
\newcommand{\utdmhad}{UTD-MHAD}
\newcommand{\mmact}{MMACT}
\newcommand{\mmactimpro}{12.37\% (F1-Measure)}
\newcommand{\approachname}{Fusion-GCN}
\newcommand{\githuburl}{\url{https://github.com/mduhme/fusion-gcn}}

\newcommand{\tableutdmhad}{

        \scriptsize
        \begin{tabular}{l|r}
            \hline
            \textbf{Approach}                                                 & \textbf{Acc} \\ \hline
                    Skeleton                                                 & 92.32          
            \\ 
            \hline
            RGB Patch Features R-18     & 27.67 \\ 
            RGB Encoder R-18     & 27.21 \\ 
            R(2+1)D                                                           & 61.63                   \\ 
            Skeleton + RGB Encoder R(2+1)D          & 91.62          
            \\ 
            Skeleton + RGB Encoder R-18        & 89.83     
            \\ 
            Skeleton + RGB Patch Features R-18 & 73.49                   \\ 
            Skeleton + RGB Patch Features R-18 \scriptsize{(no MLP)}  & 44.60 \\ 
    
            \hline
            Skeleton + IMU (Center)                                     & 94.42                   
            \\ 
            Skeleton + IMU (Wrist/Hip)                                  & 94.07                   
            \\ 
            Skeleton + IMU (Center + Add. Edges)                             & 93.26                   
            \\ 
            Skeleton + IMU (Wrist/Hip + Add. Edges)                          & 93.26         
                    \\ 
            Skeleton + IMU (Channel Fusion)        & 90.29    
            \\ 
            \hline
            Skeleton + IMU + RGB Patch Features R-18 & 78.90 \\ 
            Skeleton + IMU + RGB Encoder R-18 & 92.33 \\ 
            Skeleton + IMU + RGB Encoder R(2+1)D   & 92.85 \\ 
            \hline
            PoseMap \cite{DBLP:conf/cvpr/LiuY18}                              & \textbf{94.50}                   \\ 
            Gimme Signals \cite{DBLP:conf/iros/MemmesheimerTP20}            & 93.33                   \\ 
            MCRL \cite{2019ISenJ..19.1862L}                                   & 93.02                   \\ 
            \hline
        \end{tabular}
        \label{tab:comparison}

}
\newcommand{\tablemmact}{
                    \scriptsize
            \begin{tabular}{l|Hr}
                \hline
                \textbf{Approach}                      & \textbf{Acc} & \textbf{F1-Measure} \\ 
                \hline
                Skl                               &  87.85                   & 88.65               \\ 
                Skl+Acc(W+P)+Gyo+Ori         & 84.85                   & 85.50               \\ 
                Skl+Acc(W+P)+Gyo+Ori (Add. Edges)  & 84.40                   & 84.78               \\ 
                Skl+Acc(W)  & \textbf{89.32} & 89.55 \\
                Skl+Acc(P) & 87.70 & 88.72 \\
                
                Skl+Gyo & 86.35 & 87.41 \\
                Skl+Ori & 87.65 & 88.64 \\
                Skl+Acc(W+P) & 89.30 & \textbf{89.60} \\
                \hline
                SMD \cite{DBLP:journals/corr/HintonVD15} (Acc+RGB)  & -                    & 63.89               \\ 
                MMD \cite{DBLP:conf/iccv/KongWDKTM19} (Acc+Gyo+Ori+RGB) & -                       & 64.33               \\ 
                MMAD \cite{DBLP:conf/iccv/KongWDKTM19} (Acc+Gyo+Ori+RGB)& -                       & 66.45               \\ 
                Multi-GAT \cite{islam2021multi}     &   -   & 75.24 \\ 
                SAKDN \cite{DBLP:journals/tip/LiuWLL21} & & 77.23 \\
                \hline
            \end{tabular}
            \label{tab:comparison2}

}

\newif\ifreview
\reviewfalse

\ifreview
	\usepackage{lineno}

	\linenumbers
\fi

\begin{document}


\def\SubNumber{143}

\def\GCPRTrack{Fast Review Track}

\title{\approachname: Multimodal Action Recognition using Graph Convolutional Networks}

\ifreview
\else

	\author{Michael Duhme \and
	Raphael Memmesheimer\orcidID{0000-0003-3602-754X} \and
	Dietrich Paulus\orcidID{0000-0002-2967-5277}}
	
	\authorrunning{M. Duhme et al.}
	
	\institute{Arbeitsgruppe Aktives, University of Koblenz-Landau, Koblenz, Germany\\
	\email{\{mduhme,raphael,paulus\}@uni-koblenz.de.de}}
\fi

\maketitle              

\begin{abstract}

In this paper we present \approachname{}, an approach for multimodal action recognition using \glspl{gcn}.
Action recognition methods based around \glspl{gcn} recently yielded state-of-the-art performance for skeleton-based action recognition. 
With \approachname{}, we propose to integrate various sensor data modalities into a graph that is trained using a \gls{gcn} model for multi-modal action recognition. Additional sensor measurements are incorporated into the graph representation either on a channel dimension (introducing additional node attributes) or spatial dimension (introducing new nodes).
\approachname{} was evaluated on two publicly available datasets, the \utdmhad{}- and \mmact{} datasets, and demonstrates flexible fusion of RGB sequences, inertial measurements and skeleton sequences. Our approach gets comparable results on the \utdmhad{} dataset and improves the baseline on the large-scale \mmact{} dataset by a significant margin of up to \mmactimpro{} with the fusion of skeleton estimates and accelerometer measurements.
\end{abstract}

\section{Introduction}
Automatic \gls{har} is a research area that is utilized in various fields of application where human monitoring is infeasible due to the amount of data and scenarios where quick reaction times are vital, such as surveillance and real-time monitoring of suspicious and abnormal behavior in public areas \cite{DBLP:conf/icmcs/NiuLHW04, DBLP:journals/tip/HuXFZM07, DBLP:conf/cvpr/NiYK09, DBLP:journals/air/TripathiJA18} or intelligent hospitals and healthcare sectors \cite{DBLP:journals/access/GaoXXHLAJF18, DBLP:conf/cvpr/DuongBPV05} with scenarios such as fall detection \cite{Noury2007, DBLP:conf/iccvw/SolbachT17}, monitoring of medication intake \cite{Huynh2009} and detection of other potentially life-threatening situations \cite{DBLP:conf/cvpr/DuongBPV05}. Additional areas of applications include video retrieval \cite{DBLP:journals/air/RamezaniY16}, robotics \cite{DBLP:conf/hri/RyooFXAM15}, smart home automation \cite{DBLP:journals/corr/abs-1904-10354}, autonomous vehicles \cite{Zheng2018}.
In recent years, approaches based on neural networks, especially \glspl{gcn}, like ST-GCN \cite{DBLP:conf/aaai/YanXL18} or 2s-AGCN \cite{DBLP:conf/cvpr/ShiZCL19a}, have achieved state-of-the-art results in classifying human actions from time series data. 


\begin{figure}
    \centering
    \subfloat[Appended to center joint]{
        \includegraphics[width=0.3\linewidth]{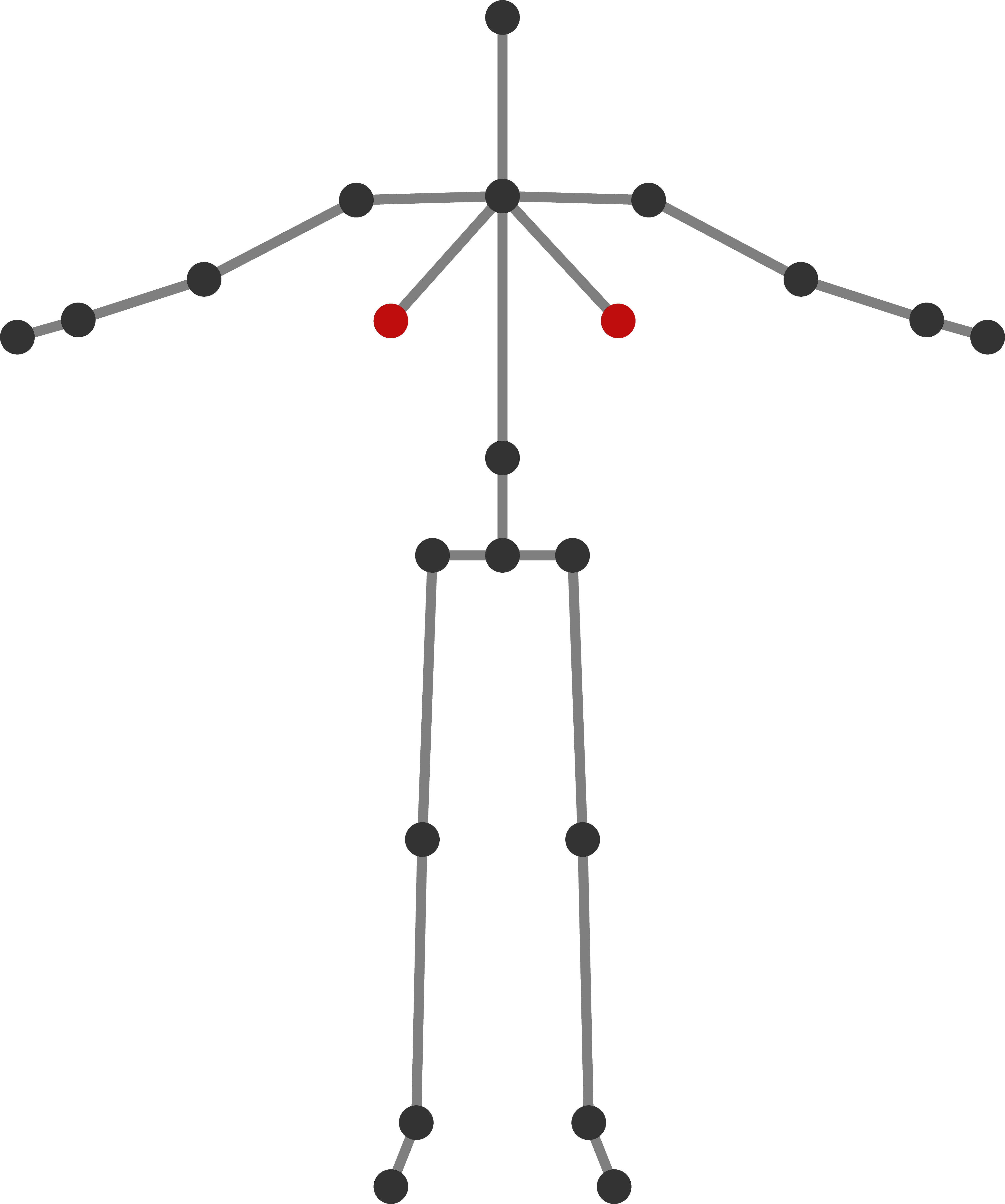}
    }
    \hspace{2em}
    \subfloat[Appended to right wrist/hip]{
        \includegraphics[width=0.3\linewidth]{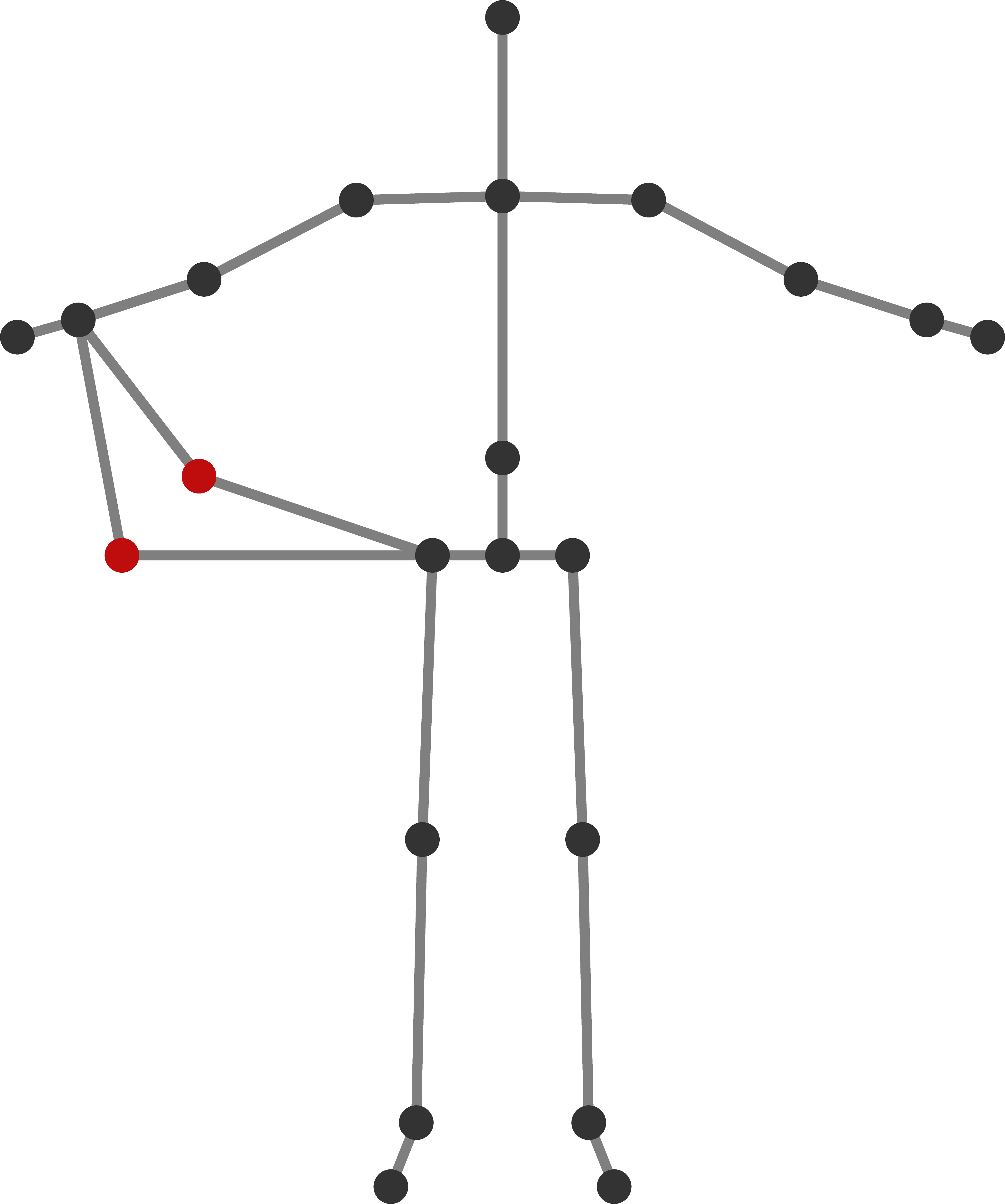}
    }
    \caption[Skeleton + IMU Graph Nodes Variations]{Showing the skeleton as included in UTD-MHAD. IMU nodes are either appended to the central node (neck) or to both the right wrist and right hip. Two additional representations arise when all newly added nodes are themselves connected by edges.}
    \label{fig:skeleton-imu-nodes}
\end{figure}

\glspl{gcn} can be seen as an extension to \glspl{cnn} that work on graph-structured data \cite{DBLP:conf/iclr/KipfW17}.
Its network layers operate by including a binary or weighted adjacency matrix, that describes the connections between each of the individual graph nodes.
As of now, due to their graph-structured representation in the form of joints (graph nodes) and bones (graph edges), research for \gls{har} using \glspl{gcn} is mostly limited to skeleton-based recognition.
However, the fusion of additional modalities into \glspl{gcn} models are currently neglected.
For that reason, taking skeleton-based action recognition as the foundation, our objective is to research possibilities of incorporating other vision-based modalities and modalities from worn sensors into existing \gls{gcn} models for skeleton-based action recognition through data fusion and augmentation of skeleton sequences.
\figname \ref{fig:skeleton-imu-nodes} gives an example of two suggestions on how inertial measurements can be incorporated into a skeleton graph. To the best of our knowledge, \approachname{} is the first approach proposing to flexibly incorporate additional sensor modalities into the skeleton graph for \gls{har}.
We evaluated our approach on two multimodal datasets, UTD-MHAD \cite{DBLP:conf/icip/ChenJK15} and \mmact{} \cite{DBLP:conf/iccv/KongWDKTM19}.

The contributions of this paper can be summarized as:
(1) We propose the fusion of multiple modalities by incorporating sensor measurements or extracted features into a graph representation. The proposed approach significantly lifts the state-of-the-art on the large-scale \mmact{} dataset.
(2) We propose modality fusion for \glspl{gcn} on two dimensionality levels: (a) the fusion at a channel dimension to incorporate additional modalities directly into the already existing skeleton nodes, (b) the fusion at a spatial dimension, to incorporate additional modalities as new nodes spatially connected to existing graph nodes.
(3) We demonstrate applicability of the flexible fusion of various modalities like skeleton, inertial, RGB data in an early fusion approach.

The code for \approachname{} to reproduce and verify our results is publicly available on \githuburl{}.
\section{Related Work}
In this section, we present related work from the skeleton-based action recognition domain that is based on \gls{gcn} and further present recent work on multimodal action recognition.
\paragraph{Skeleton-based Action Recognition}
Approaches based on \glspl{gcn} have recently shown great applicability on non-Eucliean data  \cite{DBLP:conf/aaai/PengHCZ20} like naturally graph-structure represented skeletons and have recently defined the state-of-the-art. Skeletons, as provided by large-scale datasets \cite{DBLP:journals/corr/ShahroudyLNW16}, commonly are extracted from depth cameras \cite{DBLP:series/sci/ShottonFCSFMK013}. RGB images can be transformed into human pose feature that yield a similar skeleton-graph in 2D \cite{DBLP:journals/pami/CaoHSWS21, DBLP:conf/cvpr/KreissBA19, DBLP:journals/corr/abs-1906-08172} and in 3D \cite{DBLP:journals/corr/abs-1906-08172, DBLP:journals/tog/MehtaS0XEFSRPT20, DBLP:journals/cviu/IqbalDYK0G18}. All of those approaches output skeleton-graphs that are suitable as input for our fusion approach as a base structure for the incorporation of additional modalities.
The Spatial-Temporal Graph Convolutional Network (ST-GCN) \cite{DBLP:conf/aaai/YanXL18} is one of the first proposed models for skeleton-based \gls{har} that utilizes \glspl{gcn} based on the propagation rule introduced by \acite{DBLP:conf/iclr/KipfW17}. The Adaptive Graph Convolutional Network (AGCN) \cite{DBLP:conf/cvpr/ShiZCL19a} builds on these fundamental ideas with the proposal of learning the graph topology in an end-to-end-manner.
Peng \andothers \cite{DBLP:conf/aaai/PengHCZ20} propose a \gls{nas} approach for finding neural architectures to overcome the limitations of \gls{gcn} caused by fixed graph structures.
Cai \andothers \cite{DBLP:conf/wacv/CaiJHJL21} proposes to add flow patches to handle subtle movements into a \gls{gcn}. Approaches based on \gls{gcn} \cite{DBLP:conf/cvpr/Cheng0HC0L20, DBLP:journals/corr/abs-1912-09745, DBLP:conf/mm/Song0SW20, DBLP:conf/aaai/LiLZW19} have been constantly improving the state-of-the-art on skeleton-based action recognition recently.


\paragraph{Multimodal Action Recognition}

Cheron \andothers \acite{DBLP:conf/iccv/CheronLS15} design \gls{cnn} input features based on the positions of individual skeleton joints. Here, human poses are applied to RGB images and optical flow images. The pixel coordinates that represent skeleton joints are then grouped hierarchically starting from smaller body parts, such as arms, and upper body to full body. For each group, an RGB image and optical flow patch is cropped and passed to a 2D-\gls{cnn}. The resulting feature vectors are then processed and concatenated to form a single vector, which is used to predict the corresponding action label. Similarly, \acite{DBLP:conf/ijcai/CaoZZL16} propose to fuse pose-guided features from RGB-Videos.
\acite{DBLP:journals/tcyb/CaoZZL18} further, refine this method by using different aggregation techniques and an attention model. 
\acite{DBLP:journals/corr/abs-2008-01148} propose to fuse data of RGB, skeleton and inertial sensor modalities by using a separate encoder for each modality to create a similar shaped vector representation. The different streams are fused using either summation or vector concatenation. With Multi-GAT \cite{islam2021multi} an additional message-passing graphical attention mechanism was introduced.
\acite{DBLP:journals/pr/LiXPCZS20} propose another architecture that entails skeleton-guided RGB features. For this, they employ ST-GCN to extract a skeleton feature vector and R(2+1)D \acite{DBLP:conf/cvpr/TranWTRLP18} to encode the RGB video. Both output features are fused either by concatenation or by compact bilinear correlation.

The above-mentioned multimodal action recognition approaches follow a late-fusion method, that fuse various models for each modality. This allows a flexible per modality model-design, but comes at the computational cost of the multiple streams that need to be trained.
For early fusion approaches, multiple modalities are fused on a representation level \cite{DBLP:conf/iros/MemmesheimerTP20}, reducing the training process to a single model but potentially loosing the more descriptive features from per-modality models.
\acite{DBLP:conf/iccv/KongWDKTM19} presented a multi modality distillation model. Teacher models are trained separately using a 1D-\gls{cnn}. The semantic embeddings from the teaching models are weighted with an attention mechanism and are ensembled with a soft target distillation loss into the student network. Similarly, Liu \andothers\cite{DBLP:journals/tip/LiuWLL21} utilize distilled sensor information to improve the vision modality.
Luo \andothers \cite{DBLP:conf/eccv/LuoHJNF18} propose a graph distillation method to incorporate rich privileged information from a large-scale multi-modal dataset in the source domain, and improves the learning in the target domain
More fundamentally, multimodality in neural networks is recently also tackled by the multimodal neurons that respond to photos, conceptual drawings and images of text \cite{goh2021multimodal}. Joze \andothers \cite{DBLP:conf/cvpr/JozeSIK20} propose a novel intermediate fusion scheme in addition to early and late-fusion, they share intermediate layer features between different modalities in \gls{cnn} streams. Perez-Rua \andothers \cite{DBLP:conf/cvpr/Perez-RuaVPBJ19} presented an approach for finding neural architecture search for the fusion of multiple modalities.
To the best of our knowledge, our \approachname approach is the first that proposes to incorporate additional modalities directly into the skeleton-graphs as an early fusion scheme.

\section{Approach}

In the context of multimodal action recognition, early and late fusion methods have been established to either fuse on a representation or feature level.
We present approaches for fusion of multiple modalities at representation level to create a single graph which is passed to a \gls{gcn}.

\subsection{Incorporating additional Modalities into a Graph Model}
Early fusion denotes the combination of structurally equivalent streams of data before sending them to a larger (\gls{gcn}) model, whereas late fusion combines resulting outputs of multiple neural network models. 
For early fusion, one network handles multiple data sources which are required to have near identical shape to achieve fusion. As done by Song \andothers\acite{DBLP:conf/icmcs/SongLXZ018}, each modality may be processed by some form of an encoder to attain a common structure before being fused and passed on to further networks. 
Following a skeleton-based approach, for example, by employing a well established \gls{gcn} model like ST-GCN or AGCN as the main component, RGB and inertial measurements are remodeled and factored into the skeleton structure. 
With \approachname{} we suggest the flexible integration of additional sensor modalities into a skeleton graph by either adding additional node attributes \textit{(fusion on a channel dimension)} or introducing additional nodes \textit{(fusion at a spatial dimension)}.
In detail, the exact possible fusion approach is as follows.



\begin{figure}
    \subfloat[IMU Graph Nodes]{
        \includegraphics[width=0.25\linewidth]{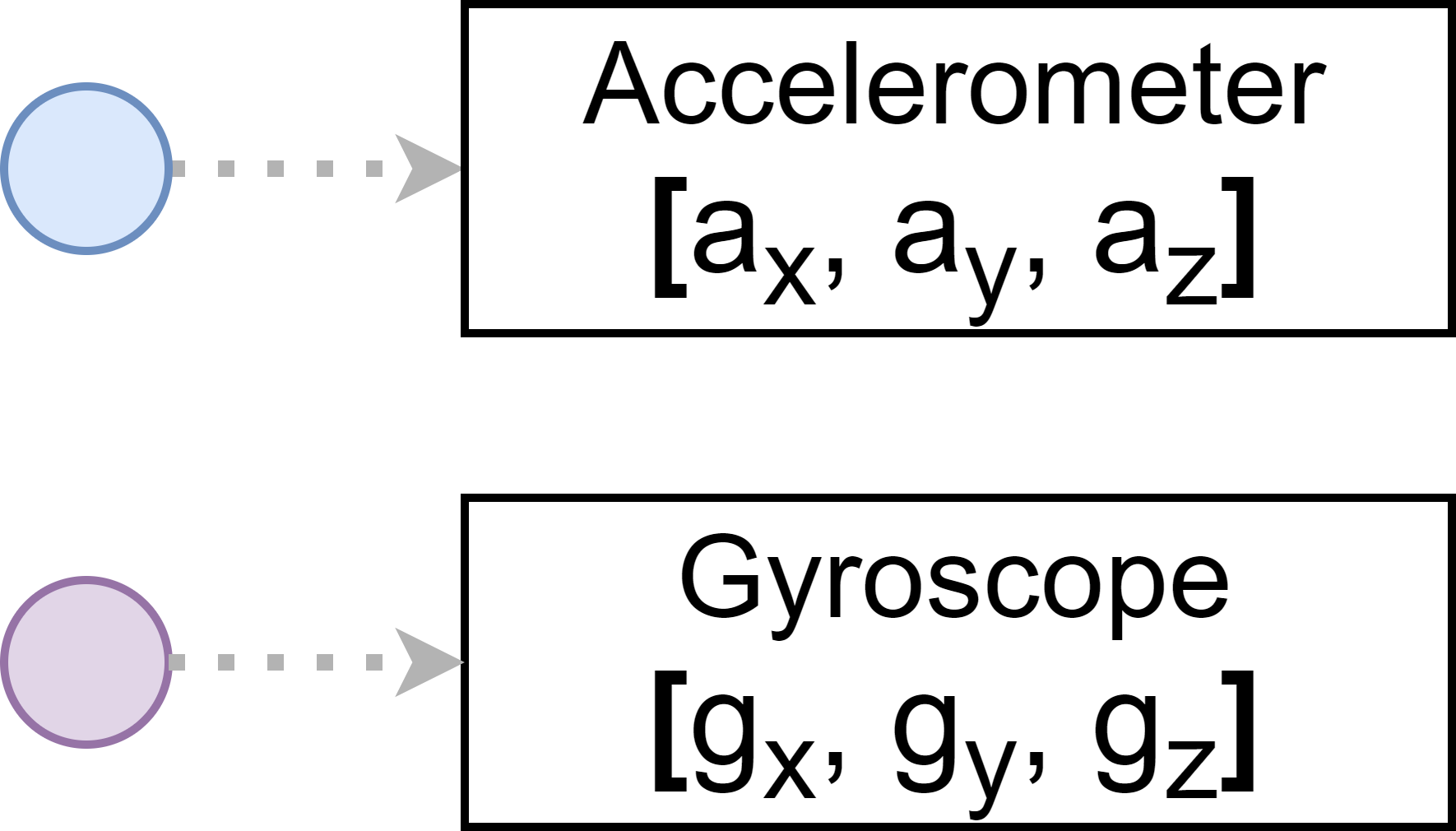}
    }
    \subfloat[Fusion at Channel Dimension]{
        \hspace{-0.5cm} 
        \includegraphics[height=0.27\linewidth]{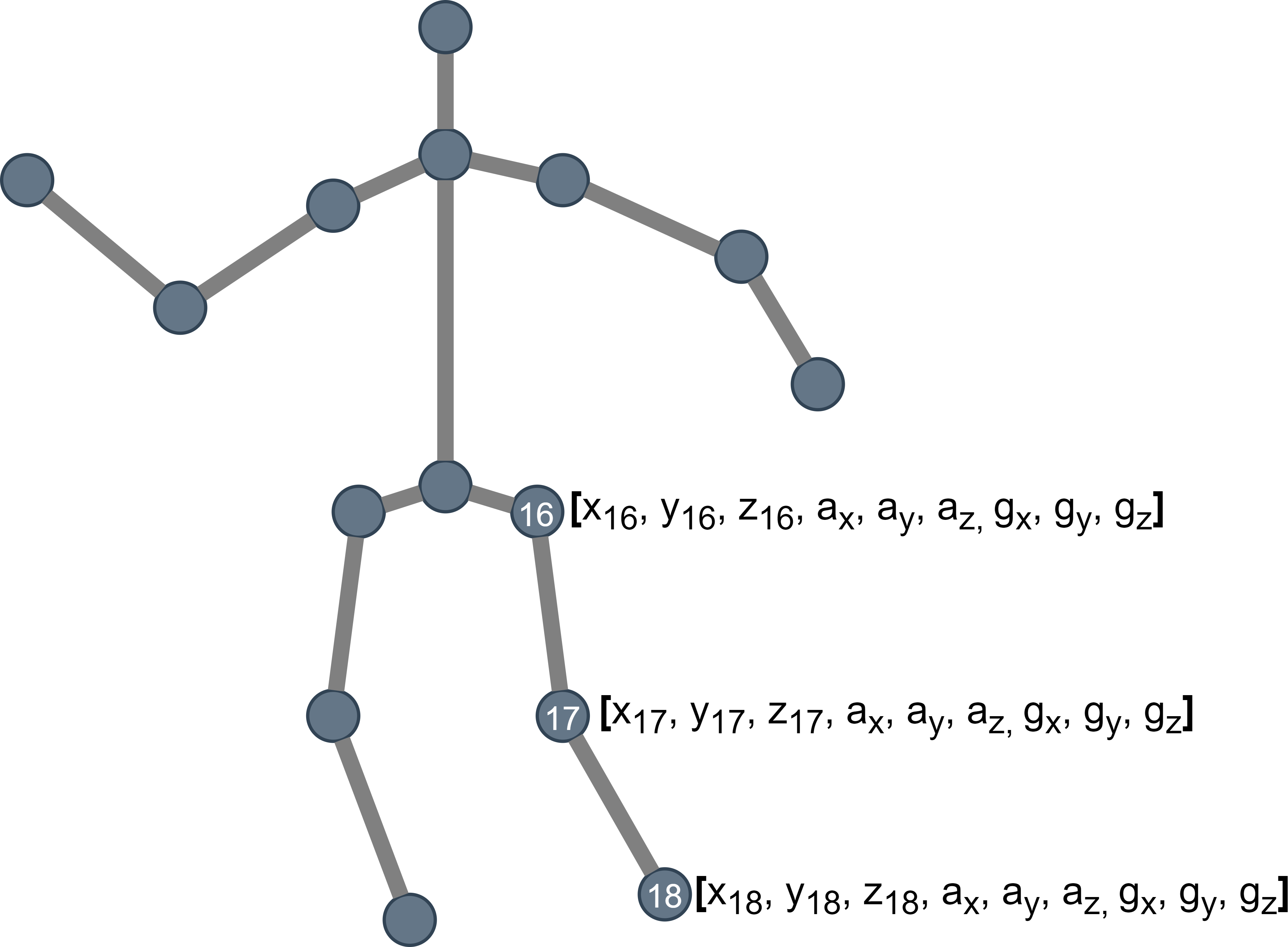}
    }
    \subfloat[Fusion at Spatial Dimension]{
        \includegraphics[height=0.27\linewidth]{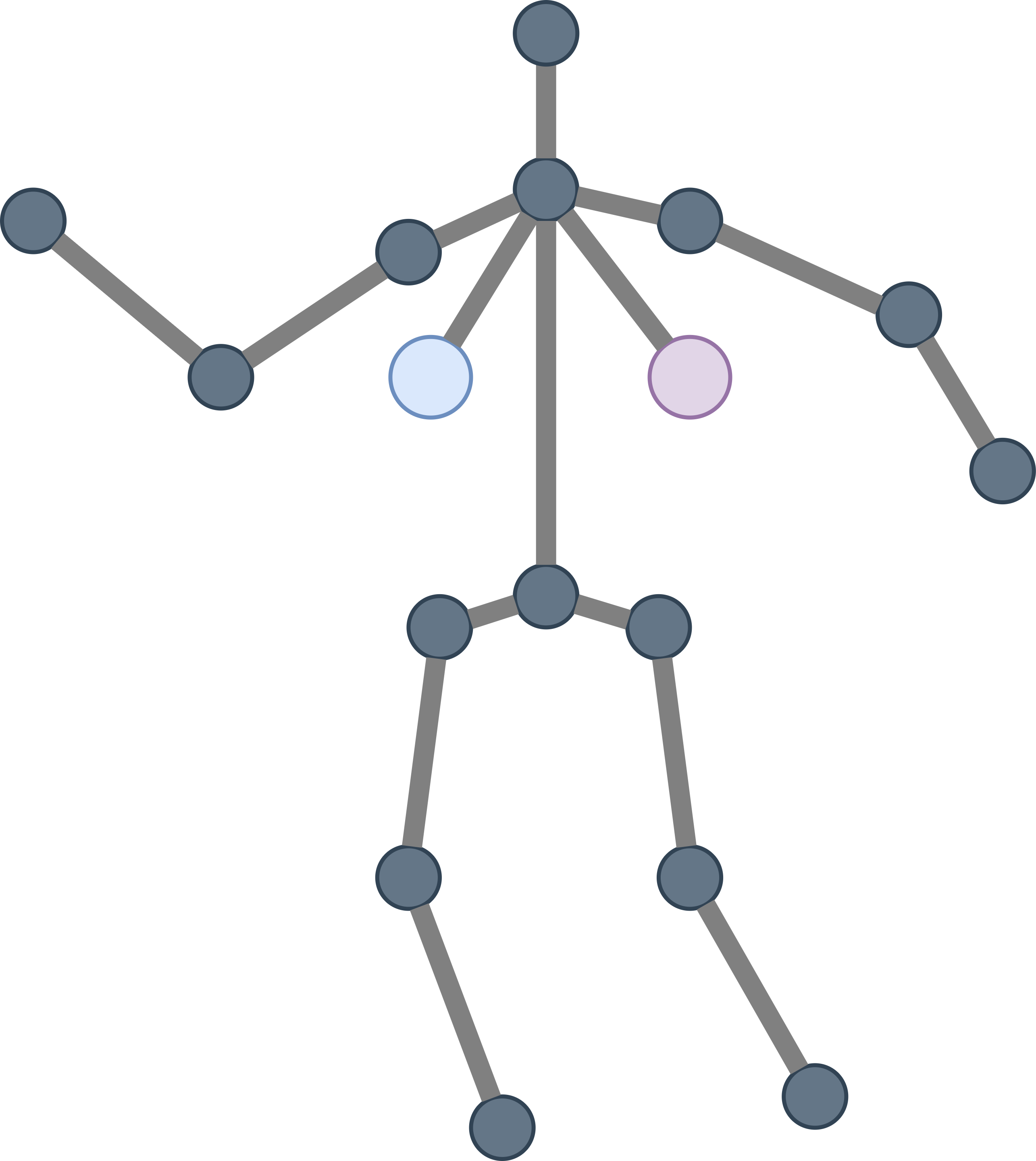}
        \hspace{0.5cm}
    }
    \caption[Fusion of Skeleton and Wearable Sensor Signals]{Options for fusion of skeleton graph and IMU signal values, viewed as skeleton nodes. If both skeleton joint coordinates and wearable sensor signals share a common channel dimension, the skeleton graph can be augmented by simply appending signal nodes at some predefined location.}
    \label{fig:skeleton-imu-fusion}
\end{figure}

Let $\mat{X}_{SK} \in \real^{(M, C_{SK}, T_{SK}, N_{SK})}$ be a skeleton sequence input, where $M$ is the number of actors that are involved in an action, $C_{SK}$ is the initial channel dimension (2D or 3D joint coordinates) and sizes $T_{SK}$ and $N_{SK}$ are sequence length and number of skeleton graph nodes. An input of shape $\real^{(M, C, T, N)}$ is required when passing data to a spatial-temporal \gls{gcn} model, such as ST-GCN.
Furthermore, let $\mat{X}_{RGB} \in \real^{(C_{RGB}, T_{RGB}, H_{RGB}, W_{RGB})}$ be the shape of an RGB video with channels $C_{RGB}$, frames $T_{RGB}$ and image size $H_{RGB} \times W_{RGB}$. For sensor data, the input is defined as $\mat{X}_{IMU} \in \real^{(M, C_{IMU}, S_{IMU}, T_{IMU})}$, where $T_{IMU}$ is the sequence length, $S_{IMU}$ is the number of sensors and $C_{IMU}$ is the channel dimension. For example, given gyroscope and accelerometer with x-, y- and z-values each, the structure would be $S_{IMU} = 2$ and $C_{IMU} = 3$. Similar to skeleton data, $M$ denotes the person wearing the sensor and its value is equivalent to that of skeleton, that is, $M_{SK} = M_{IMU}$.
Considering a multimodal model using a skeleton-based \gls{gcn} approach, early fusion can now be seen as a task of restructuring non-skeleton modalities to be similar to skeleton sequences by finding a mapping $\real^{(C_{RGB}, T_{RGB}, H_{RGB}, W_{RGB})} \to \real^{(M, C, T, N)}$ or $\real^{(M, C_{IMU}, S_{IMU}, T_{IMU})} \to \real^{(M, C, T, N)}$ with some $C$, $T$ and $N$. This problem can be reduced: If the sequence length of some modalities is different, $T_{SK} \neq T_{RGB} \neq T_{IMU}$, a common $T$ can be achieved by resampling $T_{RGB}$ and $T_{IMU}$ to be of the same length as the target modality $T_{SK}$.
Early fusion is then characterized by two variants of feature concatenation to fuse data:
\begin{enumerate}
    \setlength\itemsep{-.3em}
    \item Given $\mat{X}_{SK}$ and an embedding $\mat{X}_{E} \in \real^{(M, C_{E}, T, N_{E})}$ with sizes $C_{E}$ and $N_{E}$ where $N = N_{SK} = N_{E}$, fusion at the channel dimension
    means creating a fused feature $\mat{X} \in \real^{(M, C_{SK} + C_{E}, T, N)}$. An example is shown in Figure \ref{fig:skeleton-imu-fusion}b.
    \item Given an embedding where $C = C_{SK} = C_{E}$ instead, a second possibility is fusion at the spatial dimension, that is, creating a feature $\mat{X} \in \real^{(M, C, T, N_{SK} + N_{E})}$. Effectively, this amounts to producing $M \cdot T \cdot N_{E}$ additional graph nodes and distributing them to the existing skeleton graph at each time step by resizing its adjacency matrix and including new connections. In other words, the already existing skeleton graph is extended by multiple new nodes with an identical number of channels. 
    An example is shown in Figure \ref{fig:skeleton-imu-fusion}c.
\end{enumerate}


The following sections introduce multiple approaches for techniques about the early fusion of RGB video and IMU sensor modalities together with skeleton sequences by outlining the neural network architecture.

\subsection{Fusion of Skeleton Sequences and RGB Video}
\label{sec:rgb-fusion}

This section explores possibilities for fusion of skeleton sequences and 2D data modalities. Descriptions and the following experiments are limited to RGB video, but all introduced approaches are in the same way applicable to depth sequences.
As previously established, early fusion of RGB video and skeleton sequences in preparation for a skeleton-based \gls{gcn} model is a problem of finding a mapping $\real^{(C_{RGB}, H_{RGB}, W_{RGB})} \to \real^{(M, C, N)}$.
An initial approach uses a CNN to compute vector representations of $N \cdot M \cdot T$ skeleton-guided RGB patches that are cropped around projected skeleton joint positions.
Inspired by the work of Wang \andothers\acite{DBLP:conf/eccv/WangG18} and Norcliffe-Brown \andothers\acite{DBLP:conf/nips/Norcliffe-Brown18}, a similar approach involves using an encoder network to extract relevant features from each image of the RGB video. This way, instead of analyzing $N \cdot M \cdot T$ cropped images, the $T$ images of each video are utilized in their entirety. A \gls{cnn} is used to extract features for every frame and fuse the resulting features with the corresponding skeleton graph, before the fused data is forwarded to a \gls{gcn}. By running this procedure as part of the training process and performing fusion with skeleton sequences, the intention is to let the encoder network extract those RGB features that are relevant to the skeleton modality. For example, an action involving an object cannot be fully represented by merely the skeleton modality because an object is never part of the extracted skeleton. Objects are only visible in RGB video.

\subsection{Fusion of Skeleton Sequences and IMU Signals}
\label{sec:skeleton-imu-fusion}

Fusion of skeleton and data from wearable sensors, such as IMUs, is applicable in the same way as described in the fusion scheme from the previous section. In preparation to fuse both modalities, they again need to be adjusted to have an equal sequence length first.
Then, assuming both the skeleton joint coordinates and the signal values have a common channel dimension $C = 3$ and because $M_{SK} = M_{IMU}$, since all people wear a sensor, the only differing sizes between skeleton modality and IMU modality are $N$, the number of skeleton graph nodes, and $S$, the number of sensor signals. Leaving aside its structure, the skeleton graph is a collection of $N$ nodes. A similar understanding can be applied to the $S$ different sensors. They can be understood as a collection of $S$ graph nodes (see Figure \ref{fig:skeleton-imu-fusion}a). The fusion of sensor signals with the skeleton graph is therefore trivial because the shape is almost identical. According to channel dimension fusion as described in the previous section, the channels of all $S$ signals can be broadcasted to the x-, y- and z-values of all $N$ skeleton nodes to create the \gls{gcn} input feature $\mat{X} \in \real^{(M, (1 + S) \cdot C, T, N)}$, as presented in Figure \ref{fig:skeleton-imu-fusion}b. The alternative is to append all $S$ signal nodes onto the skeleton graph at some predefined location to create the \gls{gcn} input feature $\mat{X} \in \real^{(M, C, T, N + S)}$, as illustrated in Figure \ref{fig:skeleton-imu-fusion}c. Similar to the RGB fusion approaches, channel dimension fusion does not necessarily require both modalities to have the same dimension $C$ if vector concatenation is used. In contrast, the additional nodes are required to have the same dimension as all existing nodes if spatial dimension fusion is intended.

\begin{figure*}[btp]
    \centering
    \includegraphics[width=.85\linewidth]{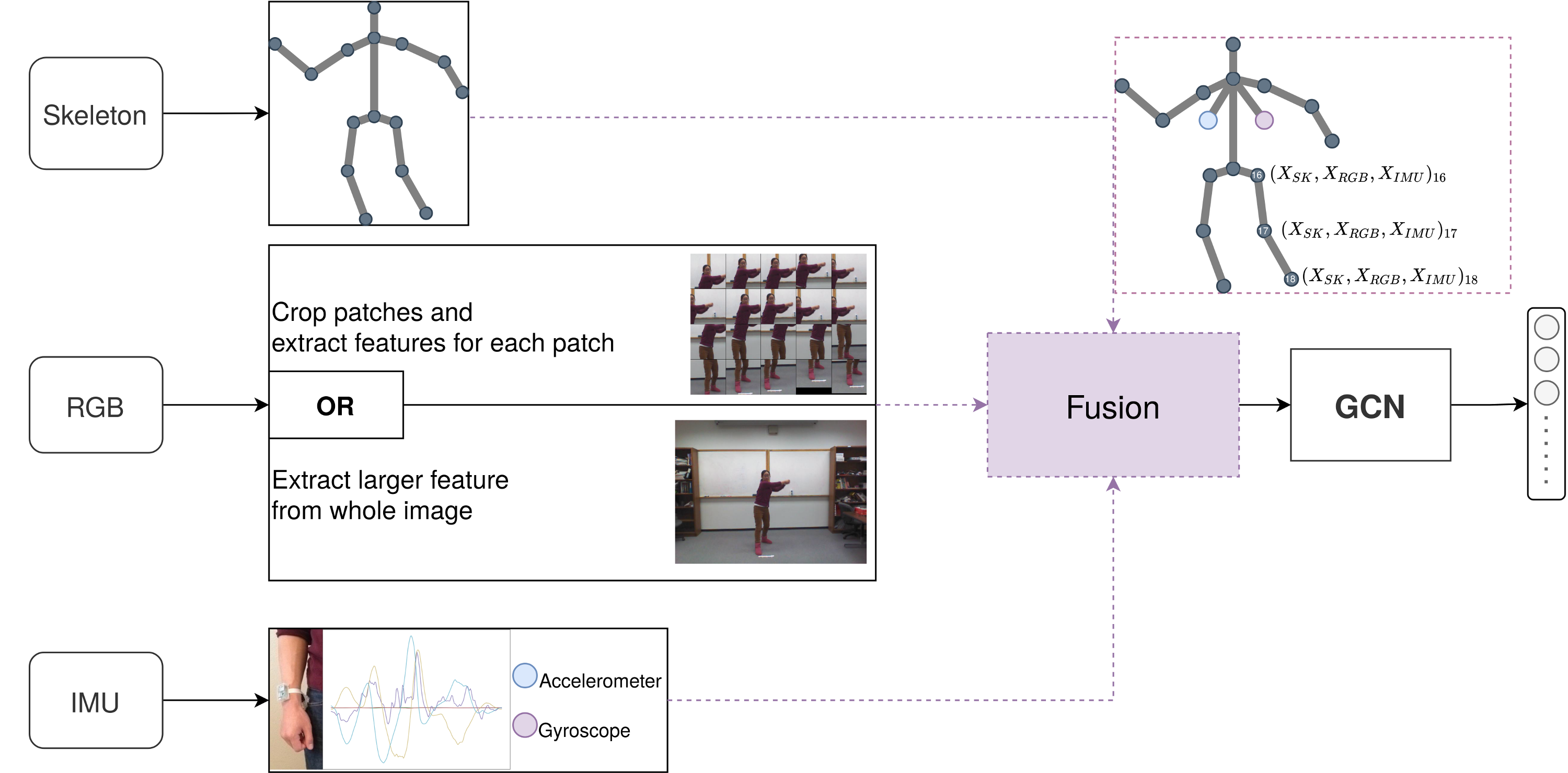}
    \caption[Combined Early Fusion Approaches]{All described approaches can be flexible fused together for early fusion and passed to a \gls{gcn}. Fusion can be realized independent of a channel or spatial fusion dimension. Here we give an example of a mixed (channel and spatial) fusion.}
    \label{fig:fusion-all}
\end{figure*}

\subsection{Combining Multiple Fusion Approaches}

All the introduced fusion approaches can be combined into a single model, as illustrated by Figure \ref{fig:fusion-all}. First, the RGB modality needs to be processed using one of the variants discussed in section \ref{sec:rgb-fusion}. Ideally, this component runs as part of the supervised training process to allow the network to adjust the RGB feature extraction process based on the interrelation of its output with the skeleton graph. Similarly, sensor signals need to be processed using one of the variants discussed previously for that modality. Assuming all sequences are identical in length, to combine the different representations, let $\mat{X}_{SK} \in \real^{(M, C_{SK}, T, N_{SK})}$ be the sequence of skeleton graphs. For RGB, let $\mat{X}_{RGB1} \in \real^{(M, C_{E}, T, N)}$ be the $C_{E}$-sized channel features obtained from computing individual patch features or feature extraction for the whole image or $\mat{X}_{RGB2} \in \real^{(M, C, T, N_{E})}$ be the RGB feature representing additional graph nodes. Respectively, the two variants of generated IMU features are $\mat{X}_{IMU1} \in \real^{(M, S \cdot C_{IMU}, T, N)}$ or $\mat{X}_{IMU2} \in \real^{(M, C_{IMU}, T, S)}$. The following possibilities to fuse different combinations of these representations arise.
\begin{itemize}
    \item $(\mat{X}_{SK}, \mat{X}_{RGB1}, \mat{X}_{IMU1}) \to \mat{X}_{FUSED} \in \real^{(M, C_{SK} + C_{E} + S \cdot C_{IMU}, T, N_{SK})}$ is the feature when combining modalities at channel dimension by vector concatenation.
    \item $(\mat{X}_{SK}, \mat{X}_{RGB1}, \mat{X}_{IMU2}) \to \mat{X}_{FUSED} \in \real^{(M, C_{SK} + C_{E}, T, N_{SK} + S)}$ combines skeleton with computed RGB features at channel dimension and expands the skeleton graph by including additional signal nodes. Since $C_{IMU} = C_{SK}$, the newly added nodes also need to be extended to have $C_{SK} + C_{E}$ channels. 
    In contrast to skeleton nodes, there exists no associated cropped patch or RGB value. Therefore, the remaining $C_{E}$ values can be filled with zeros.
    Conversely, the same applies when replacing $\mat{X}_{RGB1}$ with $\mat{X}_{RGB2}$ and $\mat{X}_{IMU2}$ with $\mat{X}_{IMU1}$.
    \item $(\mat{X}_{SK}, \mat{X}_{RGB2}, \mat{X}_{IMU2}) \to \mat{X}_{FUSED} \in \real^{(M, C, T, N_{SK} + N_{E} + S)}$ introduces new nodes for both RGB and signal modalities. This is accomplished by appending them to a specific location in the graph.
\end{itemize}

\section{Experiments}
We conducted experiments on two public available datasets and various modality fusion experiments. 
If not stated otherwise we use the top-1 accuracy as reporting metric for the final epoch of the trained model.

\subsection{Datasets}

\paragraph{\utdmhad} 

\utdmhad{} \cite{DBLP:conf/icip/ChenJK15} is a relatively small dataset containing 861 samples and 27 action classes, which thereby results in shorter training durations for neural networks. Eight individuals (four females and four males) perform each action a total of four times, captured from a front-view perspective by a single Kinect camera. 
\utdmhad{} also includes gyroscope and accelerometer modalities by letting each subject wear the inertial sensor on either the right wrist or on the right hip, depending on whether an action is primarily performed using the hands or the legs. 
For the following experiments using this dataset, the protocol from the original paper \cite{DBLP:conf/icip/ChenJK15} is used.

\paragraph{\mmact}
The \mmact{} dataset \cite{DBLP:conf/iccv/KongWDKTM19} contains more than 35k data samples and
35 available action classes. With 20 subjects and four scenes with four currently available different camera perspectives each, the dataset offers a larger variation of scenarios.
RGB videos are captured with a resolution of 1920 $\times$ 1080 pixels at a frame rate of 30 frames per second. 
For inertial sensors, acceleration, gyroscope and orientation data is obtained from a smartphone carried inside the pocket of a subject's pants. Another source for acceleration data is a smartwatch, resulting in data from four sensors in total. For the following experiments using this dataset, the protocol from the original paper \cite{DBLP:conf/iccv/KongWDKTM19} is used which proposes a cross-subject and a cross-view split. Since skeleton sequences are missing in the dataset, we create them from RGB data using OpenPose \cite{DBLP:journals/pami/CaoHSWS21}.

\subsection{Implementation}

Models are implemented using PyTorch 1.6 and trained on a Nvidia RTX 2080 GPU with 8GB of video memory. To guarantee a deterministic and reproducible behavior, all training procedures are initialized with a fixed random seed. Unless stated otherwise, experiments regarding UTD-MHAD use a cosine annealing learning rate scheduler \cite{DBLP:conf/iclr/LoshchilovH17} with a total of 60 epochs, warm restarts after 20 and 40 epochs, an initial learning rate of $1\mathrm{e}{-3}$ and ADAM \cite{DBLP:journals/corr/KingmaB14} optimization. Experiments using RGB data instead run for 50 epochs without warm restarts. Training for MMAct adopts the hyperparameters used by Shi \andothers\cite{DBLP:conf/cvpr/ShiZCL19a}.
For the MMACT, skeleton and RGB features were extracted for every third frame for more efficient pre-processing and training. 
The base GCN model is a single-stream AGCN for all experiments.

\subsection{Comparison to the State-of-the-Art}

\begin{table}[h]
    
    \centering
    \subfloat[UTD-MHAD]{
        \tableutdmhad
    }
    \subfloat[MMAct]{
        \tablemmact
    }
    \caption{Comparison to the State-of-the-Art}
\end{table}
\paragraph{\utdmhad{}}
Table \ref{tab:comparison} shows a ranking of all conducted experiments in comparison with other recent state-of-the-art techniques that implement multimodal \gls{har} on \utdmhad{} with the proposed cross-subject protocol. Without \glspl{gcn} and all perform better than the default skeleton-only approach using a single-stream AGCN. 
Additionally, another benchmark using \glspl{gcn} on \utdmhad{} does not exist, thus, making a direct comparison of different approaches difficult. From the listing in Table \ref{tab:comparison}, it is clear that all fusion approaches skeleton and IMU modalities achieve the highest classification performance out of all methods introduced in this work. 
In comparison to the best performing fusion approach of skeleton with IMU nodes appended at its central node.
MCRL \cite{2019ISenJ..19.1862L} uses a fusion of skeleton, depth and RGB to reach 93.02\% (-1.4\%) validation accuracy on \utdmhad{}. Gimme Signals \cite{DBLP:conf/iros/MemmesheimerTP20} reach 93.33\% (-1.09\%) using a CNN and augmented image representations of skeleton sequences. PoseMap \cite{DBLP:conf/cvpr/LiuY18} achieves 94.5\% (+0.08\%) accuracy using pose heatmaps generated from RGB videos. This method slightly outperforms the proposed fusion approach.

\paragraph{\mmact{}}

To show better generalization, we also conducted experiments on the large-scale \mmact{} dataset which contains more modalities, classes and samples as the \utdmhad{} dataset. Note we only use the cross-subject protocol, the signal modalities can not be seperated by view.
A comparison of approaches regarding the MMAct dataset is given in Table \ref{tab:comparison2}. Kong \andothers\acite{DBLP:conf/iccv/KongWDKTM19} propose along with the \mmact{} dataset the MMAD approach, a multimodal distillation method utiliziting an attention mechanism that incorporates acceleration, gyroscope, orientation and RGB. For evaluation, they use the F1-measure and reach an average of 66.45\%. Without the attention mechanism, the approach (MMD) yields 64.33\%. An approach utilizing the standard distillation approach Single Modality Distillation (SMD) yields 63.89\%. The current baseline is set by SAKDN \cite{DBLP:journals/tip/LiuWLL21} which distills sensor information to enhance action recognition for the vision modality. 
Experiments show that the skeleton-based approach can be further improved by fusion with just the acceleration data to reach a recognition F1-measure of 89.60\% (+12.37\%). The \mmact{} dataset contains two accelerometers, where only the one from the smartwatch yields a mention-able improvement.
The most significant improvement of our proposed approach is yielded by introducing the skeleton graph. 
In contrast, while the fusion approaches of skeleton and all four sensors do not improve the purely skeleton-based approach of 88.65\% (+13.41\%), with 85.5\% (+10.26\%) without additional edges and 84.78\% (+9.54\%) with additional edges, both reach a higher F1-measure than the baseline but also impact the pure skeleton-based recognition negatively.

\subsection{Ablation Study}

\paragraph{Fusion of Skeleton and RGB}

\begin{figure}[!ht]
    \centering
    \subfloat[Skeleton + RGB Patch Feature Fusion]{
        \includegraphics[width=.65\linewidth]{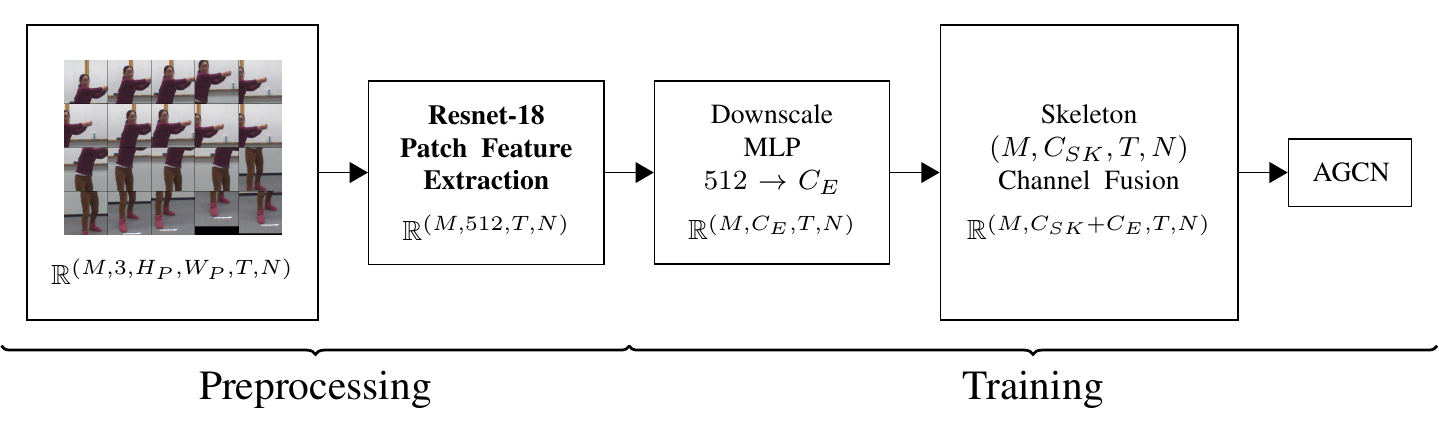}
    }

    \subfloat[Skeleton + Resnet-18 Generated Feature Fusion]{
        \includegraphics[width=.7\linewidth]{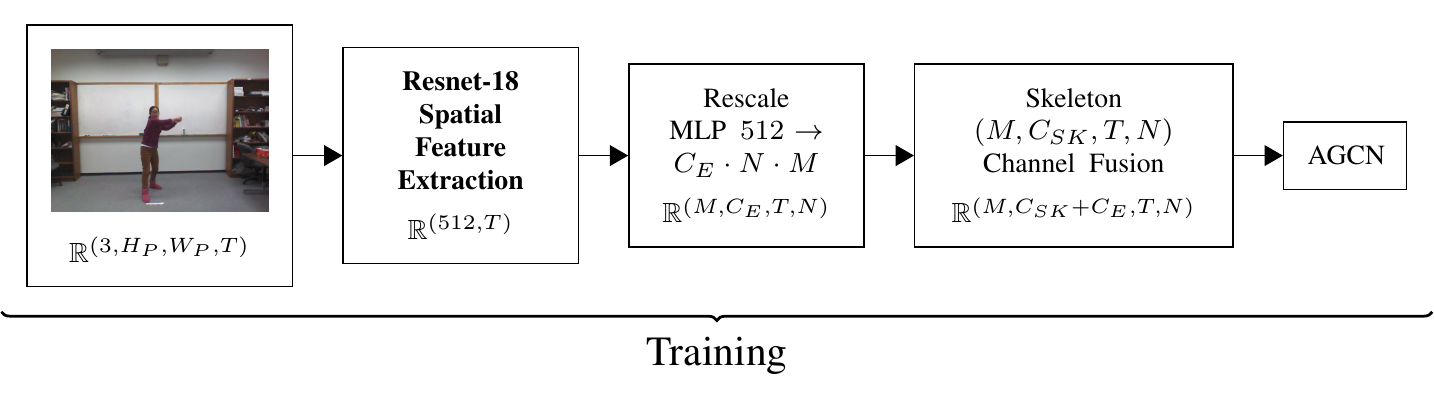}
    }

    \subfloat[Skeleton + Modified R(2+1)D Generated Feature Fusion]{
        \includegraphics[width=.7\linewidth]{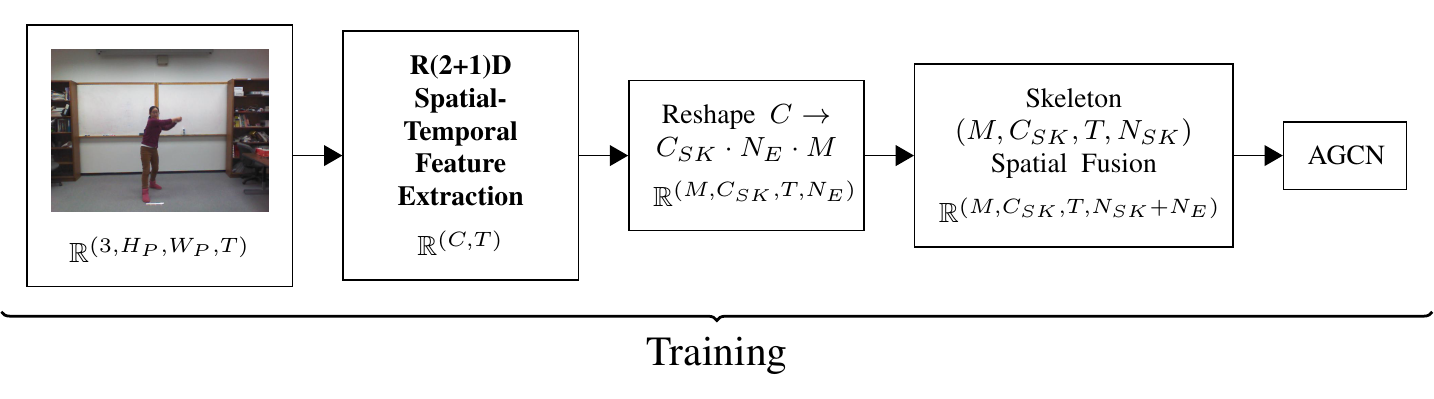}
    }
    \caption[Skeleton + RGB Fusion Models]{The three different skeleton + RGB Fusion models with reference of an image from UTD-MHAD. The first model generates a feature for each node, while the last two generate a feature for the entire image that is distributed to the nodes and adjusted as part of the supervised training.}
    \label{fig:rgb-models}
\end{figure}

Skeletons and RGB videos are combined using the three approaches depicted in \figname\ref{fig:rgb-models}. \figname\ref{fig:rgb-models}a shows an approach using RGB patches that are cropped around each skeleton node and passed to a Resnet-18 to compute a feature vector $\vec{X}_{RGB} \in \real^{(M, 512, T, N)}$ as part of preprocessing. 
The second approach, shown in \figname\ref{fig:rgb-models}b, uses Resnet-18 to compute a feature vector for each image. 
The resulting feature vector is rescaled to the size $C_E \cdot N \cdot M$ and reshaped to be able to be fused with skeleton data. Similarly, in \figname\ref{fig:rgb-models}c, the third approach uses R(2+1)D.
In terms of parameters, the basis Skeleton model has 3.454.099 parameters, only 2.532 parameters are added for incorporation of inertial measurements into the model Skeleton+IMU(Center) 3.456.631 for a 2.2\% accuracy improvement. Fusion with an RGB
encoder adds five times more parameters (Skeleton+RGB Encoder Resnet-18 with 17.868.514) and a massive training overhead.

\tabname\ref{tab:comparison} shows that the RGB approaches viewed individually (without fusion) do not reach the performance of R(2+1)D pre-trained for action recognition. 
Results regarding the fusion models show a low accuracy of 73.49\% for RGB patch features that have been created outside the training process and 44.6\% for the same procedure without a downscaling \gls{mlp}.
A similar conclusion can be drawn from the remaining two fusion models. 
Using R(2+1)D to produce features shows a slightly increased effectiveness of +1.79\% (91.62\%) over Resnet-18 (89.83\%) but -0.7\% in comparison to the solely skeleton-based approach.

\paragraph{Fusion of Skeleton and IMU}

Fusion of skeletal and inertial sensor data is done according to \figname\ref{fig:skeleton-imu-fusion}. \figname\ref{fig:skeleton-imu-nodes} shows the skeleton structure of UTD-MHAD and illustrates two possibilities for fusing the red IMU graph nodes with the skeleton by connecting them to different skeleton joint nodes. In Figure \ref{fig:skeleton-imu-nodes}a, nodes are appended at the central skeleton joint as it is defined in ST-GCN and AGCN papers. The configuration depicted in Figure \ref{fig:skeleton-imu-nodes}b is attributed to the way sensors are worn by subjects of UTD-MHAD. This configuration is therefore not used for MMAct. 
Additional configurations arise when additional edges are drawn between the newly added nodes. According to Figure \ref{fig:skeleton-imu-fusion}b, another experiment involves broadcasting the $\vecspace{\real}{6}$-sized IMU feature vector to each skeleton joint and fuse them at channel dimension.
\begin{wrapfigure}[13]{r}{0.5\textwidth}
    \centering
    \includegraphics[width=\linewidth]{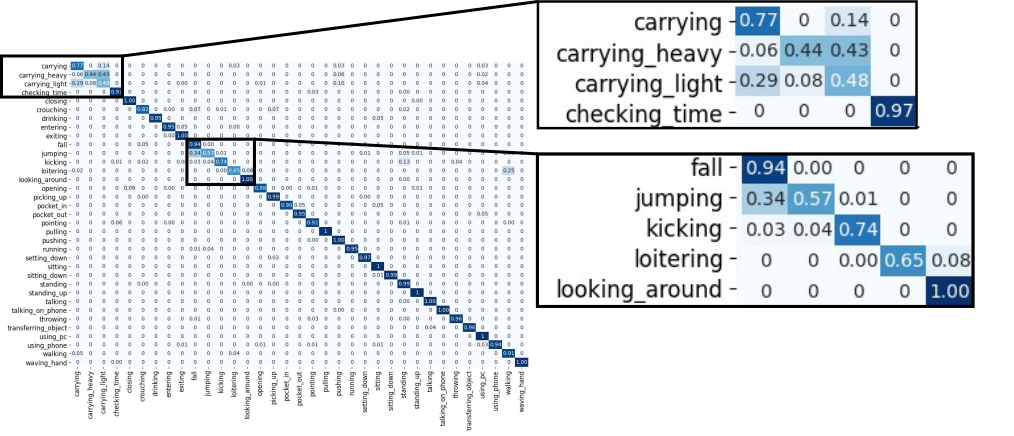}
    \caption{Confusion matrix for the results on \mmact{} with the fusion of skeleton and accelerometer measurements from the smartwatch with highlighted high-confused actions.}
    \label{fig:mmact_confusion}
\end{wrapfigure}
From the results in \tabname\ref{tab:comparison}, it is observable that all skeleton graphs with additional associated IMU nodes at each point in time improve the classification performance by at least one percent. In comparison to a skeleton-only approach, variants with additional edges between the newly added nodes perform generally worse than their not-connected counterparts and are both at 93.26\% (+0.94\%). The average classification accuracy of both other variants reaches 94.42\% (+2.1\%) and 94.07\% (+1.75\%). Despite having a slightly increased accuracy for appending new nodes to the existing central node, both variants almost reach equal performance and the location where nodes are appended seemingly does not matter much. While all experiments with fusion at spatial dimension show increased accuracies, the only experiment that does not surpass the skeleton-based approach is about fusion of both modalities at channel dimension, reaching 90.29\% (-2.03\%) accuracy.

\begin{figure}[ht]
    \centering
    \includegraphics[width=.9\linewidth]{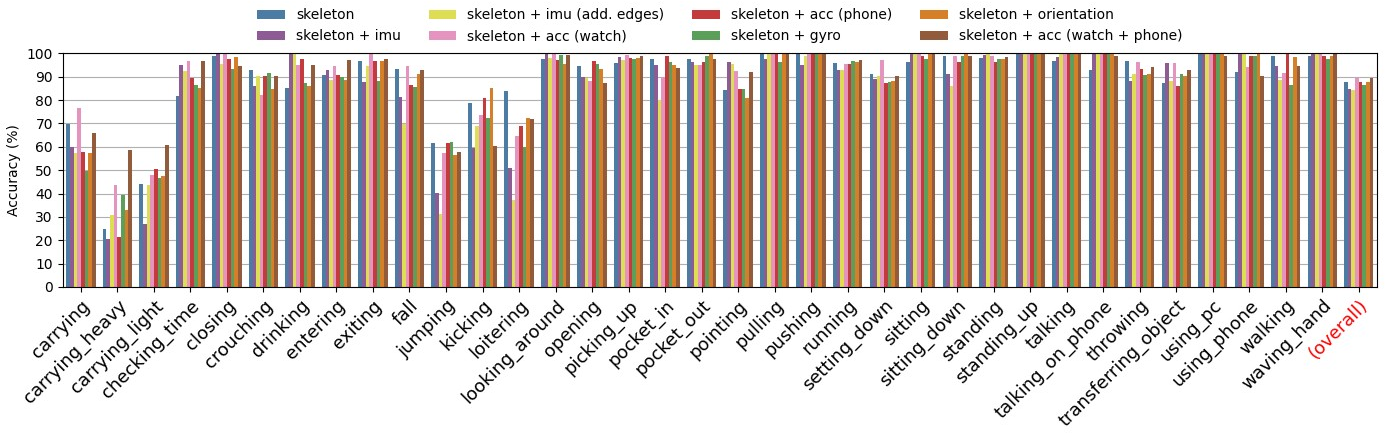}
    \caption{Class specific accuracy for all \mmact{} classes for the fusion of various data modalities with \approachname{}.}
    \label{fig:mmact_barchart}
\end{figure}
For MMAct, all experiments are conducted using only the configuration in Figure \ref{fig:skeleton-imu-nodes}a and its variation with interconnected nodes. \tabname\ref{tab:comparison2} shows that the skeleton-based approach reaches 87.85\% accuracy for a cross-subject split, fusion approaches including all four sensors perform worse and reach only 84.85\% (-3\%) and 84.4\% (-3.45\%). Mixed results are achieved when individual sensors are not part of the fusion model. Fusion using only one of the phone's individual sensors, acceleration, gyroscope or orientation, reaches comparable results with 87.70\% (-0.15\%), 86.35\% (-1.5\%) and 87.65\% (-0.2\%) accuracy, respectively. On the contrary, performing a fusion of skeleton and acceleration data obtained by the smartwatch or with the fusion of both acceleration sensors shows an improved accuracy of 89.32\% (1.47\%) and 89.30\% (1.45\%).

\begin{wraptable}[10]{r}{0.32\textwidth}
    \centering
    \scriptsize
    \vspace{-3.0\baselineskip}
    \begin{tabular}{l|r|r}
    \hline
        \textbf{Class} & \textbf{Skl} &	\textbf{Skl + Acc} \\
        \hline
        carrying\_heavy &	24.69 & 43.83 \\
        checking\_time &	81.93 & 96.58 \\ 
        drinking &	85.00 & 95.00 \\
        transferring\_object &	87.23	& 96.10 \\
        pointing &	84.52 & 92.34 \\
        \hline
    \end{tabular}
    \caption{Top-5 most improved classes by the fusion of skeleton (Skl) and additional accelerometer (Acc) data from the smartwatch.}
    \label{tab:top_5_watch_improvement}
\end{wraptable}
\tabname \ref{tab:top_5_watch_improvement} shows the top-5 improved classes by the fusion with the accelerometer measurements of a smartwatch. All the top-5 improved actions have a high arm movement in common. In \figname \ref{fig:mmact_confusion} we give a confusion matrix for the Skeleton + Accelerometer (Watch) and highlight the most confused classes. Especially the variations of the "carrying" actions are hard to distinguish by their obvious similarity. Also, actions that contain sudden movements with high acceleration peaks are often confused ("jumping" is often considered as "falling"). In general, most of the activities can be recognized quite well.
\figname \ref{fig:mmact_barchart} gives a general comparison of all class-specific results on different fusion experiments. Especially the fusion from skeleton-sequences with the accelerometer measurements (skeleton + acc (watch)) suggest a high improvement on many classes, especially the similar "carrying" classes.

\paragraph{Fusion of Skeleton, RGB and IMU}

%

One experiment is conducted using skeleton, RGB and IMU with IMU nodes appended to the skeleton central node without additional edges in combination with and all three RGB early fusion approaches.
The results in \tabname\ref{tab:comparison} show that, like previously except for the RGB patch feature model, all models achieve an accuracy over 90\%, albeit not reaching the same values as the skeleton and IMU fusion approach.



\subsection{Limitations and Discussion}

Comparing skeleton and skeleton + IMU, the fused approach generally has less misclassifications in all areas. Especially similar actions, such as "throw", "catch", "knock" or "tennis swing", are able to be classified more confidently. The only action with decreased recognition accuracy using the fused approach is "jog" which is misclassified more often as "walk", two similar actions and some of the few with sparse involvement of arm movement.
Common problems for all RGB approaches regarding UTD-MHAD are a small number of training samples, resulting in overfitting in some cases that can not be lifted by either weight decay or dropout. Another fact is the absence of object interactions in UTD-MHAD. With the exception of "sit2stand" and "stand2sit", actions such as "throwing", "catching", "pickup\_throw" or sports activities never include any objects. As pointed out previously, skeleton is focused purely on human movements and, by that, omits all other objects inside of a scene. RGB still contains such visual information, making it supposedly more efficient in recognizing object interactions. In contrast, many of \mmact{}'s actions, like "transferring\_object", "using\_pc", "using\_phone" or "carrying", make use of real objects. While fusion with RGB modality achieves similar accuracies as other approaches, incorporating the data into the network increases the training time by up to a magnitude of ten; hence, the RGB fusion models do not provide a viable alternative to skeleton and IMU regarding the current preprocessing and training configurations.
Therefore, due to timely constraints, experiments for fusion of skeleton and RGB modalities on the larger dataset \mmact{} are omitted.

\section{Conclusion}

With \approachname{}, we presented an approach for multimodal action recognition using \glspl{gcn}. To incorporate additional modalities we suggest two different fusion dimensions, either on a channel- or spatial dimension. Further integration into early- and late fusion approaches have been presented. In our experiments we considered the flexible fusion of skeleton sequences, with inertial measurements, accelerometer-, gyro-, orientation- measurements separately, as well as RGB features.
Our presented fusion approach successfully improved the previous baselines on the large-scale \mmact{} dataset by a significant margin. Further, it was showcased that additional modalities can further improve recognition from skeleton-sequences. However, the addition of too many modalities decreased the performance.
We believe that \approachname{} demonstrated successfully that \glspl{gcn} serve as good basis for multimodal action recognition and could potentially guide future research in this domain.


%
%
%
\bibliographystyle{splncs04}
\bibliography{references}

\end{document}